\documentclass[lettersize,journal]{IEEEtran}
\usepackage{amsmath,amsfonts}
\usepackage{array}
\usepackage{textcomp}
\usepackage{stfloats}
\usepackage{url}
\usepackage{verbatim}
\usepackage{graphicx}
\usepackage{cite}
\usepackage{caption}
\usepackage{graphicx}
\usepackage{subcaption}
\usepackage{adjustbox}
\usepackage{multirow}
\usepackage{algorithm}
\usepackage{algpseudocode}
\usepackage{hyperref}
\begin{document}

\title{Graph Convolutional Network With Pattern-Spatial Interactive and Regional Awareness for Traffic Forecasting}

\author{Xinyu Ji, Chengcheng Yan, Jibiao Yuan, Fiefie Zhao
        % <-this % stops a space
\thanks{This work was supported in part by the Young Innovative Talent Program of Shihezi University under Grant CXPY202207. (Corresponding author: Fiefie Zhao.)}% <-this % stops a space
\thanks{X. Ji is with the College of Mechanical and Electrical Engineering, Shihezi University, Shihezi 832003, China. (e-mail: 20211012426@stu.shzu.edu.cn)}
\thanks{C. Yan and J. Yuan are with the School of Mathematics and Computational Science, Xiangtan University, Xiangtan 411105, China. (e-mail: ycc956176796@gmail.com)(e-mail: yuan2112020@163.com)}
\thanks{F. Zhao is with the College of Sciences, Shihezi University, Shihezi, 832003, China, and also with the School of Mathematics and Computational Science, Xiangtan University, Xiangtan 411105, China. (e-mail: zhaofeifei@shzu.edu.cn) }
}

\markboth{Journal of \LaTeX\ Class Files,~Vol.~14, No.~8, August~2021}%
{Shell \MakeLowercase{\textit{et al.}}: Graph Convolutional Network With Pattern-Spatial Interactive and Regional Awareness for Traffic Forecasting}

% \IEEEpubid{1558-0016 © 2024 IEEE. Personal use is permitted, but republication/redistribution requires IEEE permission.
% }
% \IEEEpubid{\\See https://www.ieee.org/publications/rights/index.html for more information.}

\maketitle

\begin{abstract}
Traffic forecasting is significant for urban traffic management, intelligent route planning, and real-time flow monitoring. Recent advances in spatial-temporal models have markedly improved the modeling of intricate spatial-temporal correlations for traffic forecasting. Unfortunately, most previous studies have encountered challenges in effectively modeling spatial-temporal correlations across various perceptual perspectives, which have neglected the interactive fusion between traffic patterns and spatial correlations. Additionally, constrained by spatial heterogeneity, most studies fail to consider distinct regional heterogeneity during message-passing. To overcome these limitations, we propose a Pattern-Spatial Interactive and Regional Awareness Graph Convolutional Network (PSIRAGCN) for traffic forecasting. Specifically, we propose a pattern-spatial interactive fusion framework composed of pattern and spatial modules. This framework aims to capture patterns and spatial correlations by adopting a perception perspective from the global to the local level and facilitating mutual utilization with positive feedback. In the spatial module, we designed a graph convolutional network based on message-passing. The network is designed to leverage a regional characteristics bank to reconstruct data-driven message-passing with regional awareness. Reconstructed message passing can reveal the regional heterogeneity between nodes in the traffic network. Extensive experiments on three real-world traffic datasets demonstrate that PSIRAGCN outperforms the State-of-the-art baseline while balancing computational costs.
\end{abstract}


\section{Introduction}
\IEEEPARstart{W}{ith} the help of available massive urban traffic data collected from sensors on the road, cabs, private car trajectories, and transaction records of public transportation, big traffic data analysis has become an indispensable part of smart city development for traffic planning, control, and condition assessment \cite{1,2}. Traffic forecasting, which aims to predict urban dynamics using observed historical traffic data, is critical for traffic services such as flow control, route planning, and flow detection. Accurate traffic forecasting can reduce road congestion, facilitate city traffic network management, and improve transportation efficiency \cite{3}.

\begin{figure}[!htbp]
    \centering
    % 上图
    \begin{subfigure}[b]{\linewidth}
        \centering
        \includegraphics[width=\linewidth]{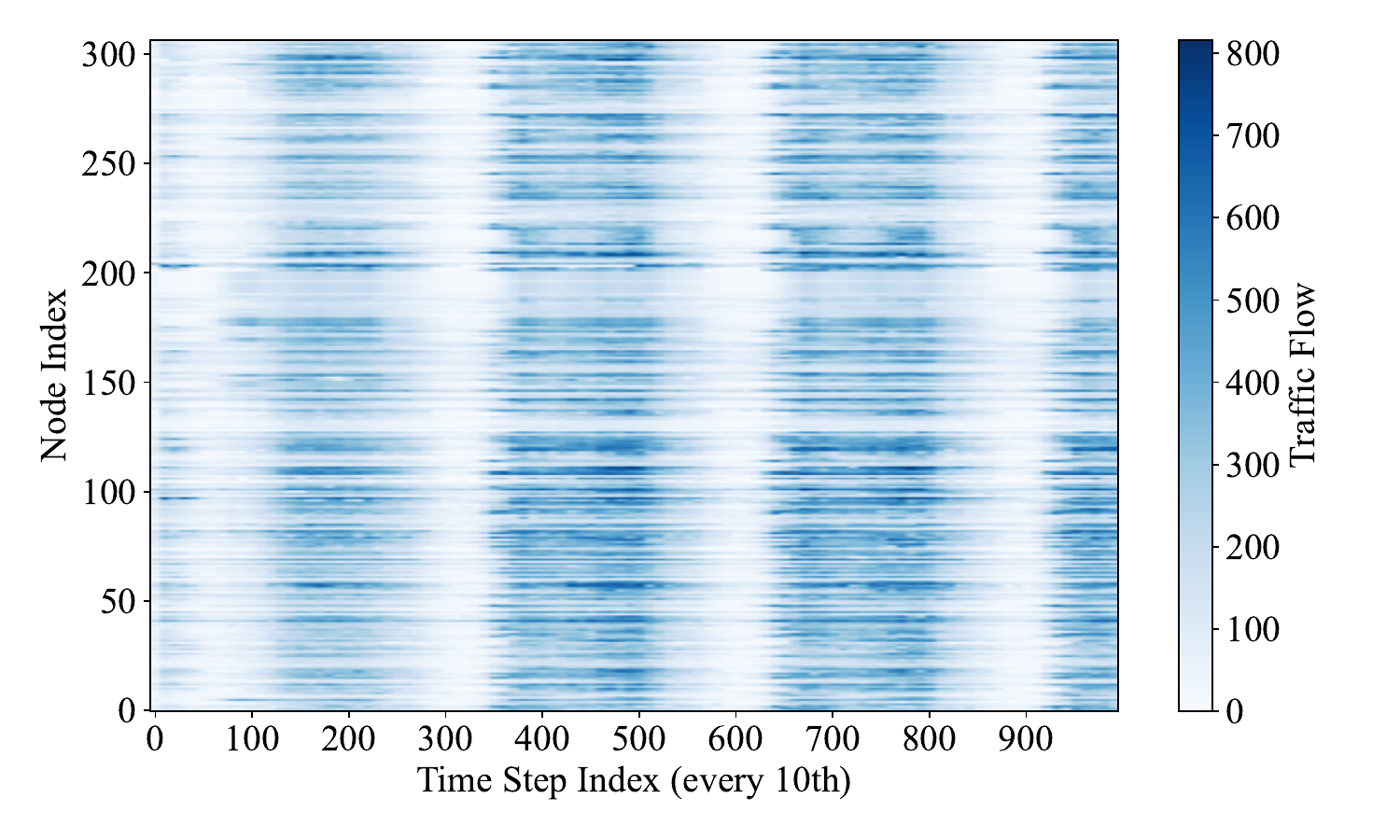}
        \captionsetup{font=footnotesize, skip=2pt}
        \caption{Distinct Pattern correlations of traffic data.}
        \label{f1:top}
    \end{subfigure}
    \vspace{4pt} % 增加上下图间距
    % 下图
    \begin{subfigure}[b]{\linewidth}
        \centering
        \includegraphics[width=\linewidth]{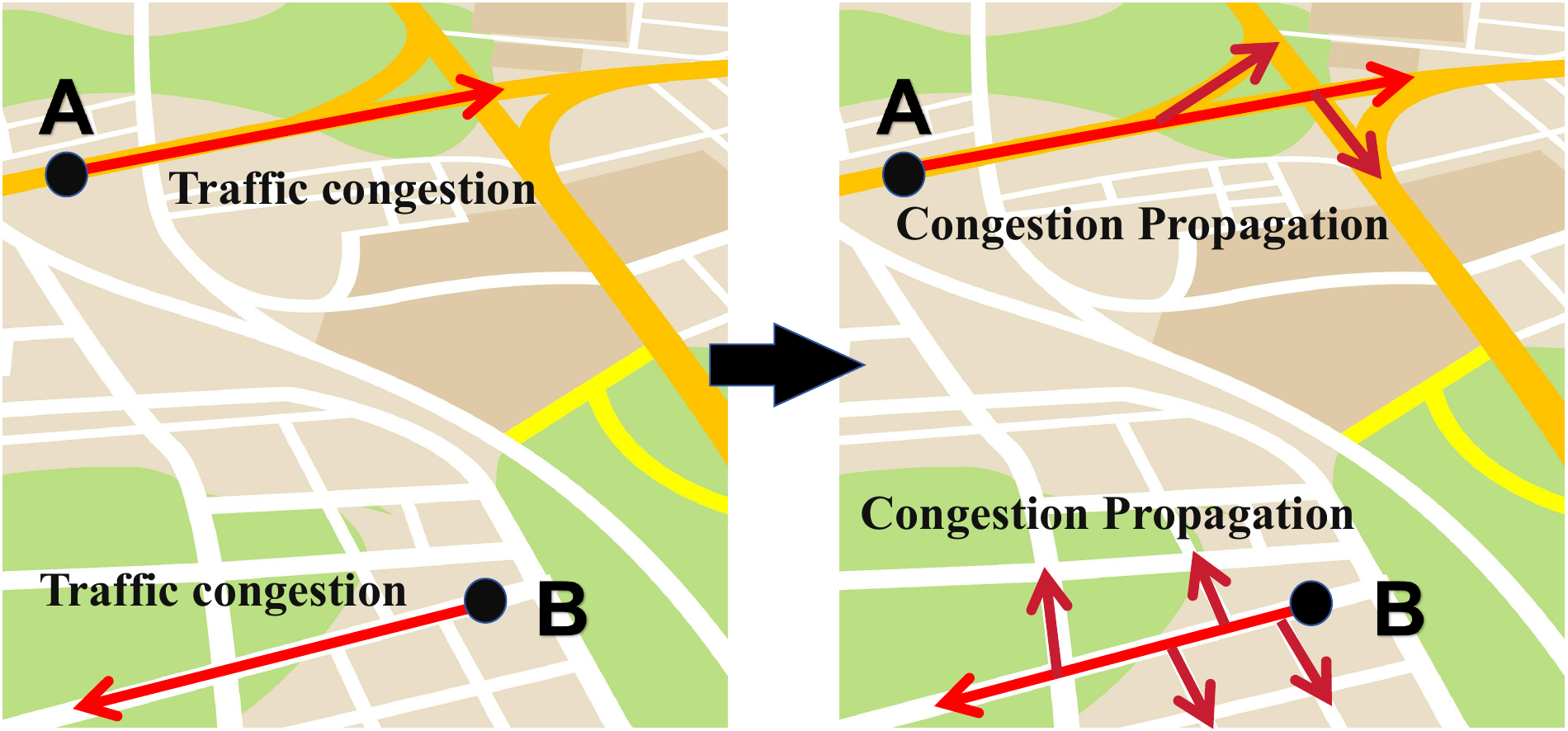}
        \captionsetup{font=footnotesize, skip=2pt}
        \caption{Regional heterogeneity between the nodes.}
        \label{f2:bottom}
    \end{subfigure}
    \captionsetup{font=footnotesize, skip=2pt}
    \caption{Spatial-temporal correlations of traffic data. (a) shows the distinct Pattern correlations of traffic data. (b) shows the regional heterogeneity between the nodes in the traffic network.}
    \label{fig:stacked}
\end{figure}

Traffic forecasting has been a prominent research focus for several decades. Numerous dedicated research endeavors have been undertaken to enhance the forecasting performance. Nevertheless, the field continues to face specific challenges. These challenges arise mainly from the complex temporal dependencies and dynamic spatial correlations in traffic data. First, traffic data exhibit distinct patterns as multivariate time series data. This can be illustrated in Fig. \ref{f1:top}, demonstrating the traffic flow of the entire traffic network on four consecutive days. The complex temporal correlation of traffic data also makes forecasting challenging. Furthermore, the propagation process of traffic flow in one area is noticeably different from that in other areas of the traffic network over time, which we refer to as regional heterogeneity in transportation dissemination. For instance, the impact of traffic congestion on the propagation of traffic flow varies across different areas, as shown in Fig. \ref{f2:bottom}. Finally, spatial heterogeneity and the different propagation processes between nodes in a traffic network are difficult to capture to uncover the spatial correlations of traffic data. Spatial heterogeneity means that different areas (e.g., residential areas and business districts) have different characteristics \cite{4}, such as road types, road width, and POIs. All of the above make traffic forecasting extremely challenging.

In response to these challenges, deep learning-based models are replacing traditional approaches in traffic forecasting \cite{4}. These models typically use temporal modules, such as Recurrent Neural Networks (RNNs) \cite{5,6,7}, Temporal Convolution Networks (TCNs) \cite{8,9,10}, and attention mechanisms \cite{11,12,13} to model temporal correlations in traffic data. Additionally, Graph Neural Networks (GNNs), such as Graph Convolutional Networks (GCNs) \cite{14,15,16} and Graph Attention Networks (GATs) \cite{17,18,19} are used as spatial modules for capturing spatial correlations in traffic data. Despite significant achievements in modeling spatial-temporal correlations, two limitations of existing models persist.

\textit{A. Failing to Effectively Consider Distinct Traffic Patterns Due to Temporal and Spatial Heterogeneity When Constructing Models}

 Some studies have employed a sequential \cite{14,15,20} or parallel \cite{11,17,21} approach by combining spatial and temporal modules and capturing spatial and temporal correlations separately. This represents a coarse-grained spatial-temporal modeling paradigm that treats traffic signals across all input time steps as a whole. This approach may weaken valuable features, amplify unimportant ones, and lacks modeling of spatial-temporal patterns. Other studies \cite{6, 7, 22} aimed to maximize the preservation of the coupling between spatial and temporal correlations by embedding spatial modules into temporal modules, e.g., embedding a GCN into an RNN. This constitutes a fine-grained spatial-temporal modeling paradigm that separately captures the temporal and spatial correlations of traffic signals at each time step. Nevertheless, it lacks awareness of global traffic patterns, making it challenging to capture spatial-temporal dependencies. Additionally, effectively modeling traffic patterns contributes to the model’s ability to reveal spatial associations within a traffic network, and vice versa. Thus, interactive fusion between pattern and spatial representations can be mutually reinforcing and yield positive feedback. However, most studies simply fuse the captured spatial and temporal correlations, ignoring the advantages of interactive learning.

\textit{B. Failing to Model Spatial-Temporal Correlations From Multiple Perspectives and Neglect the Regional Heterogeneity in Message-passing}

STGNNs are required to construct graph structures to facilitate message passing within traffic networks, thereby enabling information aggregation among nodes. Graph structures constructed using existing methods can be categorized into three types: predefined, adaptive, and dynamic. Predefined graph structures are constructed based on a priori knowledge (e.g., geographic adjacencies or distances between nodes) \cite{6,8,14} reflecting only superficial spatial relationships between nodes and incapable of modeling their latent spatial associations. The adaptive graph structure parameterizes the spatial associations between nodes by defining trainable embeddings \cite{9,23,24}, enabling the adaptive learning of latent spatial associations between nodes. However, both the predefined and adaptive graph structures cannot describe regional heterogeneity between nodes because they do not adjust to regional changes during message passing. Recent research \cite{19,22,25} has focused on building data-driven dynamic graph structures. In these studies, graph structures are adjusted in response to changes in the input spatial-temporal data, thereby enhancing their ability to capture evolving spatial correlations over time. Although existing data-driven dynamic graph construction methods have been successful, these studies have failed to consider the distinct regional heterogeneity exhibited by each node owing to spatial heterogeneity. The traffic flow at each node is affected not only by the real-time traffic flow in other areas but also by regional heterogeneity, such as special traffic patterns resulting from regional or road differences.

To address the first limitation, we propose a pattern-spatial interactive fusion strategy that models the pattern-spatial correlations by capturing changing perspectives from global to local. This strategy not only preserves the capturing capability of local representations but also captures long-range pattern-spatial dependencies. Specifically, we introduce the Pattern-Spatial Interaction (PSI) module, which is composed of pattern and spatial modules. The PSI module divides traffic signals into two classes as a pattern library and interactive fusion objects, utilizing pattern’s periodicity and trend characteristics. During the pattern-spatial interactive fusion process, the captured pattern correlations improve the ability to capture spatial correlations. This facilitates the capture of deeper pattern correlations and creates a positive feedback mechanism.

To address the second limitation, we propose a Regional Awareness Graph Convolutional Network (RAGCN) module based on message passing. In the RAGCN module, we propose a regional characteristics bank for storing distinct regional characteristics at each node owing to spatial heterogeneity. Through a data-driven approach, the RAGCN module calculates the message-passing weight with the regional characteristics bank to construct message passing. The generated message-passing process with regional awareness enhances the capture of spatial correlations.

In the end, the model proposed for the limitations above is named the \textit{\textbf{\underline{P}}attern-\textbf{\underline{S}}patial \textbf{\underline{I}}nteractive and \textbf{\underline{R}}egional \textbf{\underline{A}}wareness \textbf{\underline{G}}raph \textbf{\underline{C}}onvolutional \textbf{\underline{N}}etwork} (PSIRAGCN). The contributions of this study are summarized as follows:
\begin{itemize}
    \item A pattern-spatial interactive fusion strategy is designed to model spatial correlations from multiple perspectives. Interactive fusion between pattern and spatial correlations forms a positive feedback mechanism for the representation mining.
    \item A data-driven dynamic message-passing based approach is designed to model regional associations in traffic networks. Message passing fully considers the distinct regional heterogeneity of each node.
    \item Extensive experiments are conducted on three real-world datasets from previous work. The experimental results show that PSIRAGCN has better performance while balancing the computational costs compared with the baseline models.
\end{itemize}

\section{Related work}
In earlier times, classical statistical models were used for traffic forecasting, such as Historical Averages (HA), AutoRegressive Integrated Moving Average (ARIMA) \cite{26}, and Vector Auto-Regressive (VAR) \cite{27}. However, these are linear time-series-based models that rely on static assumptions. Since traffic data are complex nonlinear data, these models naturally underperform compared to machine learning-based models. To capture complex nonlinear relationships in traffic data, some traditional machine learning models are applied to traffic forecasting, such as Support Vector Regression (SVR) \cite{28}, Random Forest Regression (RFR) \cite{29}, and K-Nearest Neighbor (KNN) \cite{30}. These models are more effective but require specific experience to design manual features.

Deep learning-based models are effective for automatically capturing features for representation learning. Early machine learning models used RNN-based models (including LSTM and GRU) \cite{18,31} to capture the temporal correlations of traffic data. RNN-based models have limitations, such as error accumulation, slow training, and inability to handle long sequences. Convolutional neural networks (CNNs), on the other hand, process data in parallel and require relatively low memory usage. Consequently, some CNN-based models are widely used for time series \cite{32,33}. Recently, SCINet \cite{34} expanded the receptive field of convolutional operations and achieved multi-resolution analysis in a downsample-convolveinteract manner. However, the abovementioned models do not consider the spatial dimension when modeling traffic data. To address this limitation and capture spatial correlations within traffic data, some studies have used CNNs for spatial correlation capture. In these studies, CNN treat traffic data as Euclidean data, i.e., the traffic network is divided into grids to predict traffic conditions within each grid \cite{35,36,37}. Practically, the real-world traffic networks possess unique topologies, rendering traffic data essentially non-Euclidean. Therefore, the spatial correlations captured by CNN-based models are restricted.

Recently, spatial-temporal graph neural networks (STGNNs) have been widely used to capture the spatial correlations of traffic data. Typically, these models incorporate RNN-based \cite{6,7,22}, CNN-based \cite{8,9,21}, and attention mechanism-based \cite{11,12,38} temporal modules to capture temporal correlations in spatial-temporal modeling. GNN-based \cite{14,46} and attention mechanism-based \cite{11,13,19} spatial modules are used to capture spatial correlations, which can model non-Euclidean data and are suitable for capturing spatial correlations. As shown in Table \ref{t1}, we count representative STGNNs, categorize them according to their learned graph relation States, and list the models they employ in the spatial and temporal modules.

\begin{table}[!htbp]
\centering
\captionsetup{font=footnotesize, justification=centering, skip=1pt, belowskip=-6pt}
\caption{\\CLASSIFICATION OF SPATIAL-TEMPORAL MODELING MODELS}
\label{t1}
\renewcommand{\arraystretch}{1.5}
\begin{adjustbox}{width=\columnwidth} 
\begin{tabular}{c|c|c|c|c}
\hline
\multirow{2}{*}{Model} & Publication & Temporal & Spatial & Graph \\
                                & (Venue)     & Module   & Module  & Relation  \\ \hline
\hline
DCRNN \cite{6}     & ICLR 2018 & RNN           & GNN         & Static \\ 
STGCN \cite{14}    & IJCAI 2018 & CNN          & GNN         & Static \\ 
ASTGCN \cite{38}   & AAAI 2019 & CNN+Attn     & GNN+Attn    & Static \\ 
GWN \cite{9}       & IJCAI 2019 & TCN          & GCN         & Static \\ 
MTGNN \cite{10}    & KDD 2020  & TCN          & GCN         & Static \\ 
GMAN \cite{11}     & AAAI 2020 & Attention    & Attention   & Static \\ 
STSGCN \cite{39}   & AAAI 2020 & GCN          & GCN         & Static \\ 
AGCRN \cite{23}    & NeurIPS 2020 & RNN       & GCN         & Static \\ 
STGODE \cite{40}   & KDD 2021  & TCN          & ODE         & Static \\ 
STFGNN \cite{21}   & AAAI 2021 & CNN          & GCN         & Static \\ 
STG-NCDE \cite{41} & AAAI 2022 & NCDE         & GCN+NCDE    & Static \\ 
STJGCN \cite{42}   & TKDE 2023 & TCN          & GCN         & Static \\ 
ASTGNN \cite{12}   & TKDE 2021 & Attention    & GCN+Attn    & Dynamic \\ 
DSTAGNN \cite{25}  & ICML 2022 & CNN+Attn     & GCN+Attn    & Dynamic \\ 
D²STGNN \cite{43}  & VLDB 2022 & RNN+Attn     & GCN         & Dynamic \\ 
RAHRA \cite{20}    & TITS 2022 & Attention    & GCN         & Dynamic \\ 
FSTL \cite{44}     & TITS 2022 & CNN          & GCN         & Dynamic \\ 
MVSTT \cite{17}    & TC 2022   & CNN+Attn     & GCN+Attn    & Dynamic \\ 
STTGCN \cite{45}   & TITS 2023 & CNN          & GCN         & Dynamic \\ 
TinT \cite{13}     & IF 2023   & Attention    & GCN+Attn    & Dynamic \\ 
ADCT-Net \cite{19} & IF 2023   & Attention    & Attention   & Dynamic \\ 
MegaCRN \cite{7}   & AAAI 2023 & RNN          & GCN         & Dynamic \\ \hline
\end{tabular}
\end{adjustbox}
\end{table}

The DCRNN \cite{6} and STGCN \cite{14} were among the first studies to combine spatial and temporal modules to model spatial-temporal correlations. The DCRNN models the spatial correlations of traffic data as a diffusion process on directed graphs and uses a GRU in combination with a diffusion GCN for traffic forecasting. The STGCN employs a convolution operation fully in the time dimension and uses spectral graph convolution to capture the spatial correlations of traffic data. Since urban traffic is a dynamically changing system, a predefined graph structure with a static adjacency matrix cannot represent such dynamics. To this end, models represented by the GWN \cite{9}, AGCRN \cite{23}, and STJGCN \cite{42} capture the dynamic spatial correlations by designing an adaptive adjacency matrix with a GCN. As the model training stops, the adaptive learnable matrices are fixed; therefore, they do not describe dynamic spatial correlations over time. Therefore, these predefined graph structures based on adjacency relationships and adaptive graph structures relying on trainable embeddings describe static traffic networks and struggle to adequately capture the real-time dynamic associations between nodes. To overcome this limitation, as shown in Table \ref{t1}, some data-driven dynamic graph generation methods generate local-period dynamic graph structures by utilizing input traffic data. For example, DSTAGNN \cite{25} introduces a novel dynamic spatial-temporal aware graph based on a data-driven strategy to replace the static graph used in graph convolution methods. $\text{D}^2$STGNN \cite{43} employs a novel decoupled spatial-temporal framework to separate diffusion and intrinsic traffic information and utilizes a dynamic graph learning module to learn the dynamic features of the traffic network. MegaCRN \cite{7} introduces a meta-graph learner embedded into the encoder-decoder structure to address the spatial-temporal heterogeneity and non-stationarity implied in the traffic stream.

\begin{figure*}
    \centering
        \centering        \includegraphics[width=\linewidth]{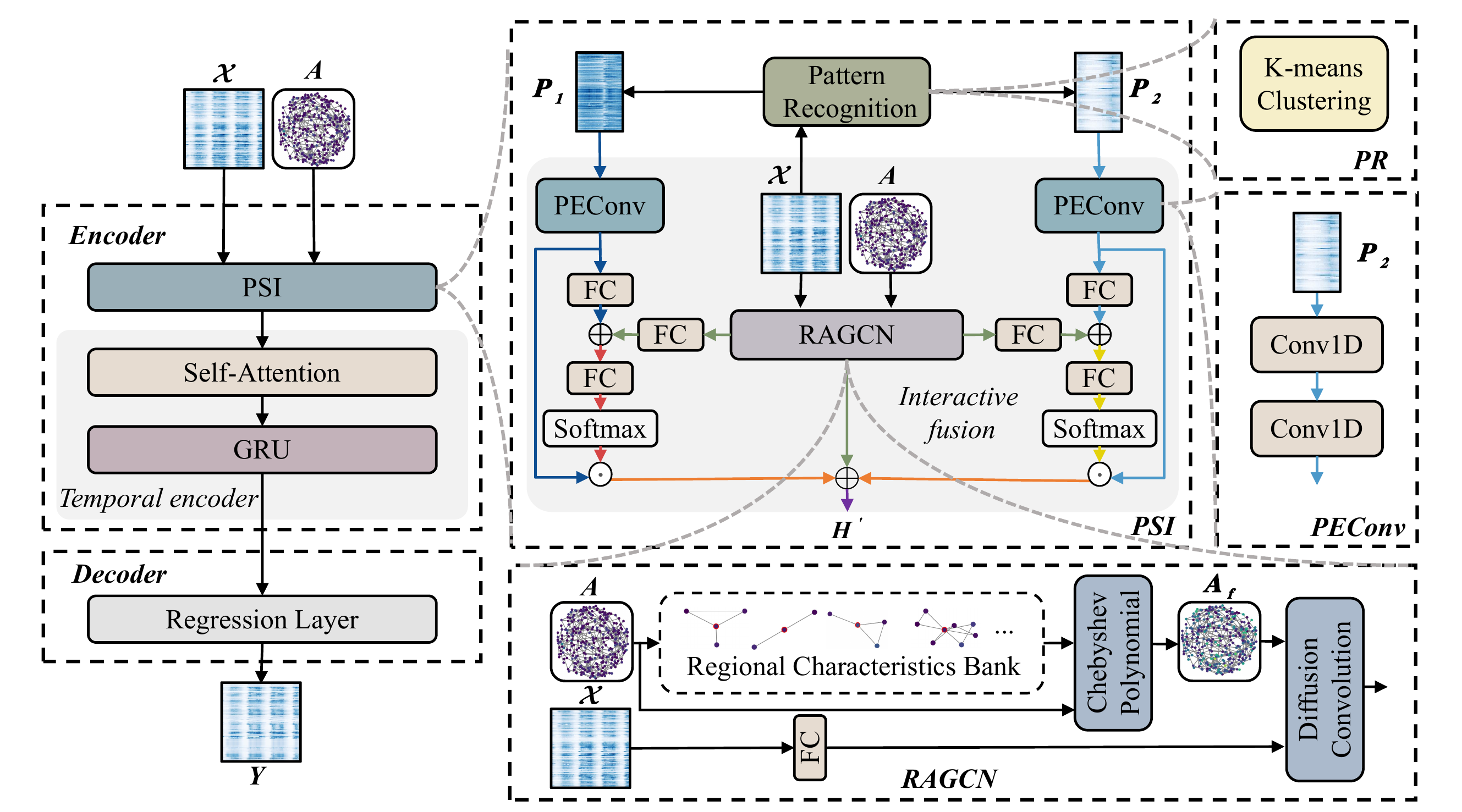}
        \captionsetup{font=small, skip=2pt}
        \caption{The framework of PSIRAGCN. The PSI module divides the input traffic signals into two patterns based on the results of the K-means clustering algorithm. The PSI module combines multiple PEConvs and an RAGCN with shared weights to enable pattern-spatial interactive fusion of pattern and spatial correlations. The RAGCN comprises a regional characteristic bank and a diffusion convolutional network on graphs. The RAGCN first performs message passing with  regional awareness and then samples for spatial representations.}
        \label{f2}
\end{figure*}

In the context of the remarkable success of transformers in natural language processing and computer vision, as illustrated in Table \ref{t1}, the latest STGNNs tend to incorporate attention mechanisms to model spatial-temporal correlations, particularly in capturing long-range dependencies. ASTGCN \cite{38} and ASTGNN \cite{12} combine the attention mechanism with GCN to model spatial-temporal correlations; the attention mechanism is mainly used to capture temporal correlations, and GCN is used to capture spatial correlations. The GMAN \cite{11} is an encoder-decoder structure consisting purely of spatial-temporal attention modules and simulates the influence of spatial-temporal factors on traffic conditions. To prevent receptive field bias, TinT \cite{13} employs an innovative mixture of long and short-range information routing mechanisms, along with an original anisotropic graph aggregation designed for unbalanced traffic flow propagation. To capture enduring spatial-temporal dependencies and unveil latent graphic feature representations that surpass temporal and spatial constraints, the ADCT-Net \cite{19} leverages an adaptive dual-graphic method incorporating a cross-fusion strategy. Although attention-based STGNNs excel in terms of effectiveness, they impose a significant computational burden.

Compared with previous studies, our proposed PSIRAGCN offers the following advantages: 
\begin{itemize}
    \item PSIRAGCN employs a pattern-spatial interactive fusion strategy, capable of capturing pattern-spatial dependencies from global to local perspectives. This perceptual horizon not only effectively captures patterns but also significantly enhances the model’s capability to capture long-range dependencies.
    \item Through our pattern-spatial interactive fusion strategy, PSIRAGCN enables captured pattern and spatial correlations to leverage each other, exploring deeper pattern-spatial representations through positive feedback.
    \item By leveraging the regional characteristics bank, PSIRAGCN can explore regional characteristics for each node and use them for massge passing. This allows the PSIRAGCN to model deeper spatial correlations between nodes while considering the distinct spatial heterogeneity of each node.
\end{itemize}

\section{PRELIMINARY}
\subsection{Definitions}
 \textit{\textbf{1) Traffic Network:}} Real-world traffic networks can be described using spatial information obtained from internal sensor networks or by identifying between stations and road segments. A traffic network can be defined as an undirected graph $\boldsymbol{G}=(\boldsymbol{V}, \boldsymbol{E}, \mathbf{A})$, where $\boldsymbol{V}$ denotes the set of $|\boldsymbol{V}|=N$ nodes, and each node denotes an observation point (sensor or road segment) in the traffic network, and $\boldsymbol{E}$ is a set of edges. $\mathbf{A} \in \mathbb{R}^{N \times N}$ is the adjacency matrix of graph $\boldsymbol{G}$, which represents the degree of association between the nodes.

 \textit{\textbf{2) Traffic signal:}} $\boldsymbol{X}^{(t)} \in \mathbb{R}^{D \times N}$ denotes the data collected by all observation points in the traffic graph $\boldsymbol{G}$ at time step $t$, where $N$ denotes the number of nodes ($N$ sensors), $D$ denotes the initial number of feature channels (e.g., the demand, volume or speed).

\subsection{Problem Formalization}
\textit{\textbf{Traffic Forecasting:}} The traffic forecasting task is to predict the future traffic signal sequence $Y=\left[\boldsymbol{Y}^{(t+1)}, \boldsymbol{Y}^{(t+2)}, \ldots,\boldsymbol{Y}^{ (t+T^{\prime})}\right]$ using a segment of historical sequence $\mathcal{X}=\left[\boldsymbol{X}^{(t-T+1)}, \boldsymbol{X}^{(t-T+2)}, \ldots, \boldsymbol{X}^{(t)}\right]$, where $T$ denotes the length of a given historical time series, and $T^{\prime}$ denotes the length of the time series to be predicted. Therefore, the traffic forecasting task can be defined as:
\begin{equation}
\left[\boldsymbol{X}^{(t-T+1)}, \ldots, \boldsymbol{X}^{(t)}\right] \xrightarrow{F}\left[\boldsymbol{Y}^{(t+1)}, \ldots, \boldsymbol{Y}^{\left(t+T^{\prime}\right)}\right],
\end{equation}
where $F$ denotes the mapping function from the historical sequence to the predicted sequence.

\section{METHODOLOGY} %% {{{
\subsection{Overview Framework} %% {{{
The framework of the proposed PSIRAGCN is illustrated in Fig. \ref{f2}. and followed an encoder-decoder architecture. The PSI encoder employs a pattern-spatial interactive fusion strategy to model the pattern-spatial correlations, and the temporal encoder employs a self-attention mechanism and GRU to model the temporal correlations. Subsequently, the decoder is utilized to regressively predict future traffic signals by integrating the captured spatial-temporal representations.

First, the raw input data are fed into the PSI module. The PSI module employs Pattern Extraction Convolution (PEConv) modules to learn complex pattern correlations. Simultaneously, we introduced an RAGCN module to capture spatial correlations and implemented a pattern-spatial interactive fusion strategy for pattern-spatial feature interactions.

As shown in Fig. \ref{f2}, the pattern-spatial interactive fusion is implemented using a divide-and-conquer approach. We divided the input traffic signals into two patterns based on the results of the K-means clustering algorithm. Next, the two patterns are passed through the PEConv to capture the pattern correlations, and the input traffic signals are passed through the RAGCN module to capture spatial correlations and perform informative interactions with the higher-dimensional representations of the patterns. In the RAGCN, a regional characteristic bank is employed to reconstruct message passing in the traffic network. This regional message passing is used to reveal deeper spatial correlations. Through the pattern-spatial interactive fusion strategy, this deeper spatial correlation is fed back into the captured pattern correlation, thereby mining more complex pattern correlations. This establishes a positive feedback pattern-spatial modeling paradigm, in which pattern and spatial correlations mutually reinforce each other.

After the PSI module was processed, pattern-spatial representations were generated. Next, these pattern-spatial representations are passed through the temporal encoder. Finally, these pattern spatial-temporal representations are passed through the regression layer, which outputs the final prediction.

\subsection{Pattern-Spatial Interaction}

Traffic data are a type of multivariate time-series data that exhibit apparent patterns along the time dimension. Patterns are crucial for exploring deep-seated spatial correlations. Inspired by previous studies \cite{34,48}, we propose the PSI module for pattern-spatial modeling, leveraging global patterns within sequences. The PSI module comprises pattern modules capturing pattern correlations, a spatial module capturing spatial correlations, and a spatial-temporal interactive strategy. This strategy involves the interaction between the pattern and spatial modules, which exchange information to facilitate the capture of pattern and spatial correlations.

As shown in Fig. \ref{f1:top} , we visualized the three-day traffic original signals from sensors in the PEMS04 dataset. The signals exhibited distinct patterns. As shown in Fig. \ref{f2}, in the PSI module, we use the above characteristics to split the input traffic signals according to the results of the K-means clustering algorithm, generating two patterns. The two patterns preserve periodicity and trend and serve as two objects for the pattern-spatial interactive fusion process. The two patterns can leverage each other’s information through continuous interaction and mutual learning.  This process maximizes the utilization of global relationships within the original signals, thereby enhancing the pattern-spatial representation mining capabilities of both the pattern and spatial modules. By integrating pattern and spatial correlations through the PSI module, the PSIRAGCN can capture pattern-spatial correlations ranging from global to local perception granularity. This pattern-spatial modeling paradigm enables the modeling of complex pattern-spatial dependencies, thereby expanding the range of application scenarios for the model. 

\textit{\textbf{1)Pattern Recognition:}} In the PSI module, we utilize the K-means clustering algorithm to classify patterns. 

The K-means algorithm uses iteration-by-iteration refinement to produce final clustering results. This algorithm employs distance metrics to assess the similarity between samples, thereby enabling the division of the sample set into $M$ categories. The mean vector of the $j^{th}$ category $P_j, j \in[1, M]$ is defined as:
\begin{equation}
\mu_j=\frac{1}{\left|P_j\right|} \sum_{x \in P_j} x,\end{equation}where $x$ denotes the elements classified within this category and the mean vector $\mu_j$ is referred to as centroid of class $P_j$. The primary objective of the K-means clustering algorithm is to identify $M$ centroids that minimize the squared error, thereby achieving effective clustering. Consequently, the objective function of the K-means clustering algorithm can be defined as: 
\begin{equation}
E=\sum_{j=1}^M \sum_{x \in P_j}\left\|x-\mu_j\right\|_2^2,
\end{equation}the smaller the squared error $E$, the greater the similarity among samples within category $P_j$. The primary procedure of the K-means algorithm is presented in Algorithm \ref{algo:kmeans}.
\begin{algorithm}[!htbp]
\caption{K-means Clustering.}
\label{algo:kmeans}
\begin{algorithmic}[1] % [1] 显示行号
\State \textbf{Input}: Sample set $\text{dataset} = \{\mathbf{X}_1, \ldots, \mathbf{X}_T\}$
\State \textbf{Output}: Cluster set $\text{clusterset} = \{P_1, \ldots, P_M\}$
\State \textbf{Initialization:}
\State \quad Randomly select $M$ samples $\{\mu_1, \ldots, \mu_M\}$ as initial centroids
\State \textbf{Repeat until convergence:}
\For{$t = 1$ \textbf{to} $T$}
    \State \quad Calculate distances $\|\mathbf{X}_t - \mu_j\|_2$ for $j = 1,\ldots,M$
    \State \quad Assign $\mathbf{X}_t$ to nearest cluster $P_j$
\EndFor
\For{$j = 1$ \textbf{to} $M$}
    \State \quad Update centroid: $\mu_j \leftarrow \dfrac{1}{|P_j|} \sum_{x \in P_j} x$
\EndFor
\end{algorithmic}
\end{algorithm}
The determination of the cluster count $M$ is crucial for optimizing model performance. An insufficient number of categories may fail to adequately capture distinctions, whereas an excessive number of categories can lead to confusion and overfitting. Therefore, we used the contour coefficient to ascertain the optimal cluster count, which can be defined as follows:

\begin{equation}
s(t)=\frac{b(t)-a(t)}{\max \{a(t), b(t)\}},
\end{equation}where $a(t)$ denotes the mean distance from the sample $\mathbf{X}_t$ to other samples within the same class, while $b(t)$ signifies the mean distance from the sample $\mathbf{X}_t$ to samples belonging to different classes. The average contour coefficient can be defined as follows: 

\begin{equation}
\bar{s}(k)=\frac{1}{T} \sum_{t=1}^T s(t),
\end{equation}

\begin{figure}
\setlength{\tabcolsep}{0.5pt} 
  \centering
  \includegraphics[width=\linewidth]{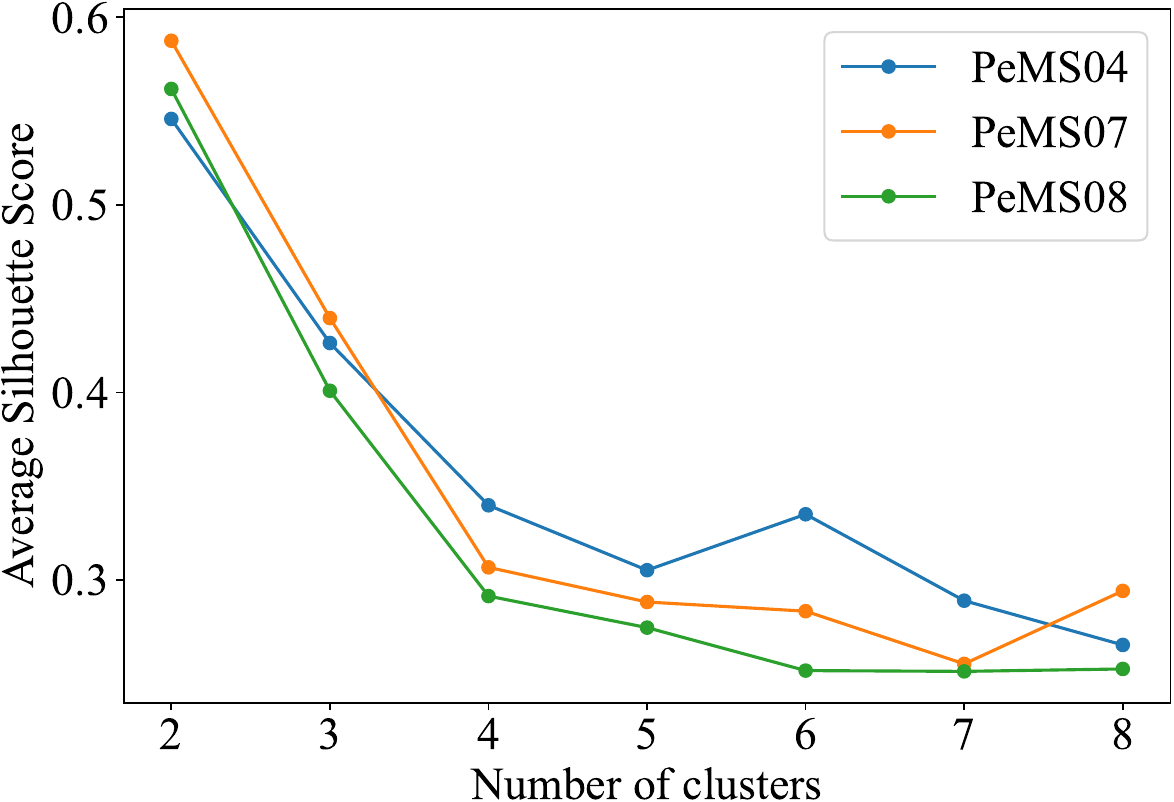}
  \caption{Average silhouette coefficients of three traffic data sets}
  \label{f4}
\end{figure}

To substantiate the rationale behind our selection of the clustering numbers, as depicted in Fig. \ref{f4}, we present the average silhouette coefficient of the pattern partition module across various datasets when employing different clustering numbers. It is evident that dividing the traffic flow data into two distinct modes is justified, as the average silhouette coefficients for the three datasets reached their peak values when the number of clusters was set to two.

\begin{figure*}
\setlength{\tabcolsep}{0.5pt} 
  \centering
\includegraphics[width=1\linewidth]{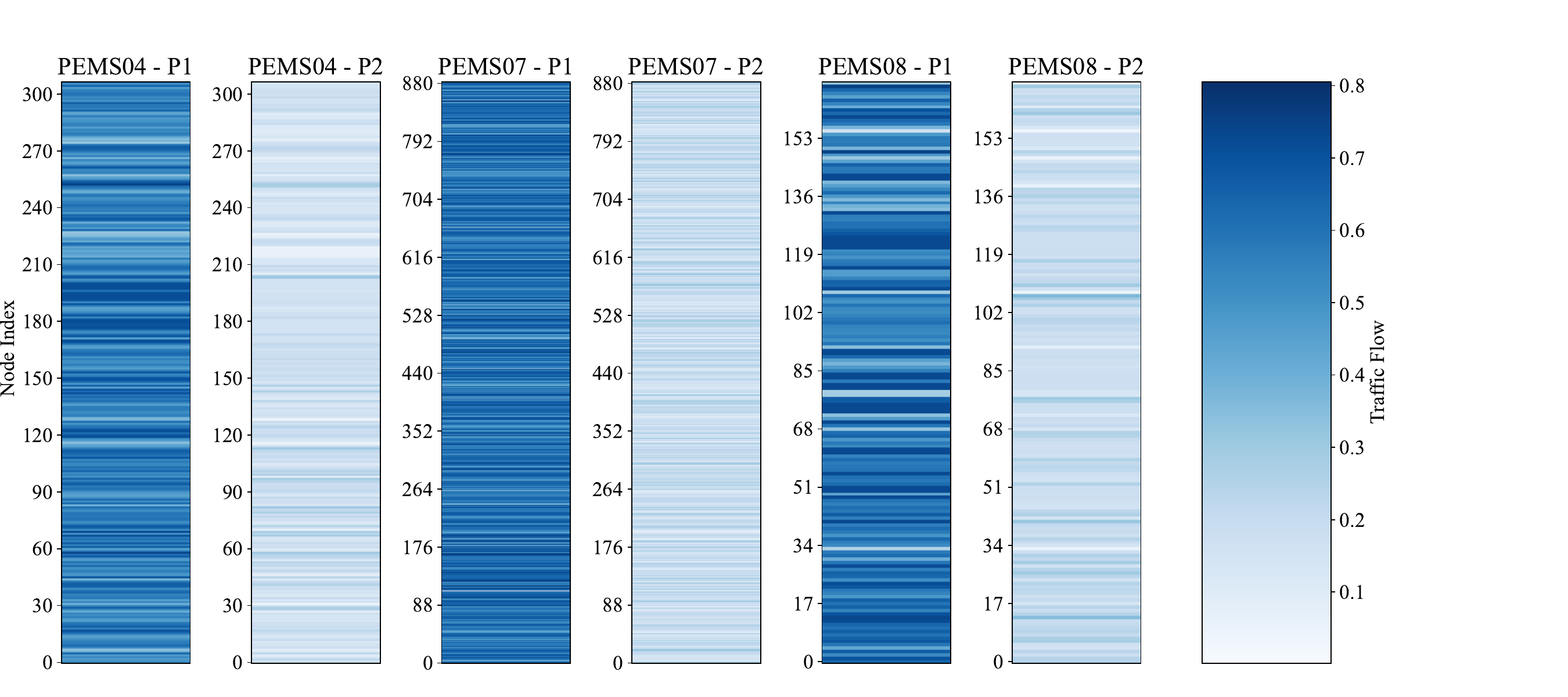}
  \caption{Clustering analysis of roads across three datasets showed distinct patterns. The longitudinal and horizontal axes represent average traffic flow and road numbers, respectively. Each dataset split into two categories: one with high traffic flow indicating peak network operation, and another with low flow suggesting low peak mode.}
  \label{f5}
\end{figure*}

Furthermore, to demonstrate the efficacy of the pattern division module, Fig. \ref{f5} shows the classification outcomes for the three datasets. Evidently, the pattern division module effectively categorizes traffic flow data into two distinct categories. Specifically, one pattern, denoted as $P_1 \in \mathbb{R}^{C \times N \times T_1}$, corresponds to the morning or evening peak periods, whereas the other pattern, $P_2 \in \mathbb{R}^{C \times N \times T_2}$, represents the low peak period of traffic flow. Here, $T_1$ and $T_2$ denote the numbers of samples within each pattern.

\textit{\textbf{2) Pattern Extraction Convolution:}} In the PSI module, we propose the PEConv module as the pattern module to capture pattern correlation. As shown in Fig. \ref{f2}, two PEConv modules are employed within the PSI module to capture pattern correlations. PEConv utilizes two layers of 1D-CNNs, each using kernel sizes of (1, $s1$) and (1, $s2$), where $s1$ and $s2$ represent the set kernel sizes. The operation in PEConv is defined as follows:
\begin{equation}
\label{eq:featmodel}
H_{P_m}=\sigma\left(\operatorname{Conv1D}\left(\sigma\left(\operatorname{Conv1D}\left(P_m\right)\right)\right)\right),\end{equation}where for $m\in\{1,2\}$ and $\sigma$ represents the sigmoid function. The correlations of the pattern are effectively and efficiently modeled in parallel through a two-layer convolution operation.

\textit{\textbf{3) Regional Awareness Graph Convolutional Network:}} We design a RAGCN module as a spatial module to capture spatial correlations. The RAGCN implements two main functions: regional awareness message passing construction, and spatial correlation capture. To fulfill these functions, the RAGCN specifically includes a regional awareness message-passing weight matrix and a diffusion-based GCN.

In contrast to previous message-passing methods, the message-passing method employed in the RAGCN considers the distinct regional heterogeneity of each node. Specifically, as shown in Fig. \ref{f2}, we define a regional characteristics bank $w_{\text{self}} \in \mathbb{R}^{N \times N}$ that can store unique regional characteristics for each node due to spatial heterogeneity during the training process. The message-passing method utilizes the regional characteristics bank $w_{\text{self}}$ to construct a message-passing weight matrix. This matrix models the regional characteristics between nodes that arise from traffic changes. The message-passing method employs a data-driven regional message-passing weight matrix construction approach. This allows message passing to be regionally adjusted based on various traffic data, ensuring that the reconstructed message passing accurately reflects the regional correlations of traffic changes.

In the message passing method, the regional characteristics bank $w_{\text {self }}^{i}$, acquired by calculating the local clustering coefficient $C C_i$ of each node through the traffic graph, can be defined as follows:
\begin{equation}
C C_i=\frac{2 E_i}{k_i\left(k_i-1\right)},  
\end{equation} 
\begin{equation}
w_{\text {self }}^{i}=1-\frac{C C_i}{\sum_{i=1}^N C C_i},
\end{equation}where $E_i$ represents the actual number of edges between the neighbors of node $i$, and $k_i$ denotes the number of neighbors, and $w_{\text {self }}^{i}$ denotes the message passing weight of the node $i$. Next, $w_{\text {self }}^{i}$ is used as a regional awareness message-passing weight in the GCN. Inspired by Chabnet \cite{chabnet}, we introduced Chebyshev polynomials to approximate the higher powers of matrices. This process generates a message-passing weight matrix with regional characteristics $A_f^k=\theta_k \mathcal{T}_k(\widetilde{L})\in \mathbb{R}^{N \times N}$, where $\theta_k$ is trainable. The generation process of the message-passing weight matrix can be defined as:
\begin{equation}
\left\{\begin{array}{l}
\mathcal{T}_0(\tilde{L})=\text{diag}(w_{\text {self }}^1,w_{\text {self }}^2,...,w_{\text {self }}^N), \\
\mathcal{T}_1(\tilde{L})=\tilde{L}, \\
\mathcal{T}_k(\tilde{L})=2 \tilde{L} \mathcal{T}_{k-1}(\tilde{L})-\mathcal{T}_{k-2}(\tilde{L}),
\end{array}\right.
\end{equation}where $\text{diag}$ represents a diagonal matrix operator, which takes as input a sequence of length $n$ and outputs an $n \times n$ matrix whose main diagonal elements are the entries of the sequence, with all other elements being zero. The scaled normalized Laplacian matrix $\tilde{L}$ is defined as $\tilde{L}=\frac{2 L}{\lambda_{\max }}-I$, and the unnormalized combined Laplacian matrix $L$ is defined as $L=I-D^{-\frac{1}{2}} A D^{-\frac{1}{2}}$, where $A \in \mathbb{R}^{N \times N}$ is an adjacency matrix, and the degree matrix $D \in \mathbb{R}^{N \times N}$ is defined as $ D = \operatorname{diag}(d_1, d_2, \dots, d_N)$, where $d_i = \sum_{j=1}^{N} A_{ij}$. The largest eigenvalue of $L$ is denoted as $\lambda_{\max}$, and $k$ represents the range of the aggregated neighborhood of the node. The regional characteristic bank $w_{\text {self }}^{i}$ can store the unique regional characteristics of each node owing to spatial heterogeneity using the local clustering coefficient. Stored regional characteristics can overcome the limitations of reconstructing graphs solely based on local period inputs. Thus, $A_f^k$  contains latent spatial associations between nodes. In contrast, $A_f^k$ performs regional characteristic calculations and yields dynamic associations between nodes. This message-passing matrix reflects the dynamic spatial associations among nodes from a regional perspective.

Finally, we used $A_f^k$ and employed a diffusion-based GCN to capture the dynamic regional spatial correlations. The diffusion-based GCN treats dynamic changes in the traffic network as diffusion processes. It aims to aggregate information about the diffusion process between nodes on the graph, where the diffusion signal at a target node depends on the recent values of its neighboring nodes. The diffusion-based GCN is defined as follows:
\begin{equation}
\begin{aligned}
H_g&=\text{RAGCN}(\mathcal{X},A_f^k)\\
&=\sum_{k=0}^{K} A_f^k \text{FC}(\mathcal{X} ),
\end{aligned}
\end{equation}where FC denotes a fully connected layer, $H_g \in \mathbb{R}^{C \times N \times T^{\prime}}$ represents the output of RAGCN.

The RAGCN first considers the regional heterogeneity of nodes owing to spatial-temporal heterogeneity as a perspective for constructing a message passing weight matrix and then utilizes this matrix to capture features, thereby modeling dynamic regional correlations. It is important to note that the PSIRAGCN performs dynamic regional message passing in the PSI module. This means that the PSIRAGCN can model dynamic  regional associations at different global and local granularities to explore deeper dynamic regional spatial correlations.

\textit{\textbf{4) Pattern-Spatial Interactive Fusion:}}
The pattern-spatial interactive fusion strategy involves pattern and spatial feature interactions between subsequences. Assuming that $\mathcal{X} \in \mathbb{R}^{C \times N \times T}$ represents the input to the PSI module. The subsequences of $\mathcal{X}$ obtained after pattern recognition (based on the K-means clustering algorithm) can be denoted as $P_1\in \mathbb{R}^{C \times N \times T_1}$ and $P_2\in \mathbb{R}^{C \times N \times T_2}$. The PEConv modules within the PSF module are denoted as PEConv1 and PEConv2. The output after the first round of interactive fusion consists of $\boldsymbol{H}_{P_1}^s \in \mathbb{R}^{C \times N \times T }$ and $\boldsymbol{H}_{P_2}^s \in \mathbb{R}^{C \times N \times T }$ (as indicated by the red and yellow arrows in Fig. \ref{f2}). These sequences, $\boldsymbol{H}_{P_1}^s$ and $\boldsymbol{H}_{P_2}^s$, undergo additional interactive fusion in the form of feedback. The resulting final subsequence is denoted as $\boldsymbol{H}_{\text {P}}^{\prime} \in \mathbb{R}^{C \times N \times T }$ (indicated by the purple arrows in Fig. \ref{f2}). The operations within the PSF module are defined as follows:
\begin{equation}
    \begin{aligned}
        P_1, P_2 &= PR(\mathcal{X}),
    \end{aligned}
\end{equation}
\begin{equation}
    \begin{aligned}
        \boldsymbol{H}_{P_1}^s &=\text{FC}(\text{FC}(\text{PEConv1}(P_1))+\text{FC}(\sigma(\text{RAGCN}(\mathcal{X},A_f^k)))),
    \end{aligned}
\end{equation}
\begin{equation}
    \begin{aligned}
        \boldsymbol{H}_{P_2}^s &=\text{FC}(\text{FC}(\text{PEConv2}(P_2))+\text{FC}(\sigma(\text{RAGCN}(\mathcal{X},A_f^k)))),
    \end{aligned}
\end{equation}
\begin{equation}
    \begin{aligned}
        \boldsymbol{a}_{P_1}^s &=\frac{\text{exp}({\text{FC}(\boldsymbol{H}_{P_1}^s)})}{\sum_{i=1}^{2} \text{exp}({\text{FC}(\boldsymbol{H}_{P_i}^s)})},
    \end{aligned}
\end{equation}
\begin{equation}
    \begin{aligned}
        \boldsymbol{a}_{P_2}^s &=\frac{\text{exp}({\text{FC}(\boldsymbol{H}_{P_2}^s)})}{\sum_{i=1}^{2} \text{exp}({\text{FC}(\boldsymbol{H}_{P_i}^s)})},
    \end{aligned}
\end{equation}
\begin{equation}
    \boldsymbol{H}^{\prime}=\boldsymbol{a}_{P_1}^s\odot\boldsymbol{H}_{P_1}^s + \boldsymbol{a}_{P_2}^s\odot\boldsymbol{H}_{P_2}^s + \sigma(\text{RAGCN}(\mathcal{X}, A_f^k)),
\end{equation}
where $\odot$ denotes the Hadamard product and $\sigma$ represents the sigmoid function. This spatial-pattern interactive fusion strategy allows subsequences to capture each other’s spatial and pattern features in a mutual feedback manner. The PSIRAGCN can learn the representations of sequences at a higher resolution by increasing the number of PSI module layers to handle longer sequence data. However, deeper layers also lead to increased memory usage. Our empirical study demonstrates that achieving excellent performance is possible with only one layer in most cases. Finally, we reshape the output of the PSI module into the original temporal shape, resulting in a complete pattern-spatial representation. This representation is then fused with the original input sequence through residual connections to obtain the output $H_e\in R^{C \times N \times T}$ of the spatial encoder.

Overall, the pattern-spatial interactive fusion strategy aims to perform pattern-spatial information sharing and mutual feedback between the sequence data. The shared spatial features explore deeper pattern features. The effective exploration of pattern features feeds back to the mining of spatial correlations, thus forming a positive feedback loop.

\subsection{Temporal Encoder}
The temporal encoder aims to model the dynamics of traffic data across the temporal dimension. As shown in Fig. \ref{f2}, the $H_e$ obtained from the PSI module are fed into a self-attention mechanism. Self-attention is a particular implementation of the attention mechanism in which the queries, keys, and values are the same sequence of symbol representations. Multi-Head SelfAttention \cite{2-19} is the most widely adopted self-attention in practice, and it enables the joint attention to information from different representation subspaces. The basic operation in the multi-head self-attention is the scaled dot-product attention, which is defined in Eq. \ref{e1}, where all the queries, keys and values are the same sequence of $H_e$, i.e., $Q = K = V$. The multi-head self-attention first linearly projects the queries, keys and values into different representation subspaces and then performs the attention function in parallel. Finally, the outputs are concatenated and further projected, resulting in the final output. Formally,
\begin{equation}
    \text{MHSelfAttention}(Q,K,V) = \oplus (\text{head}_1,...,\text{head}_h)W^o,
    \label{e1}
\end{equation}
\begin{equation}
    \text{head}_j = \text{Attention} (QW_j^Q,KW_j^K,VW_j^V),
\end{equation}where $h$ is the number of attention heads. $W_j^Q,W_j^K,W_j^V$ are projection matrices applied on $Q, K, V$ and $W^o$ is the final output projection matrix. Multi-head self-attention allows modeling the correlation of elements in sequences regardless of their distance, resulting in effectively global receptive fields. It provides a flexible method for capturing the complex dynamics of correlation in traffic data, thus enabling accurate long-term predictions.

In the temporal dimension, Recurrent Neural Networks (RNNs) excel in handling sample data characterized by temporal sequences. However, gradient vanishing and explosion constrain their effectiveness in tasks involving long-sequence data and modeling long-term dependencies. To ameliorate these problems, Gated Recurrent Unit (GRU) and Long Short Memory (LSTM) units were introduced. Compared to the LSTM model, the GRU model fuses the forgetting gate and input gate into a single update gate with fewer parameters, which reduces the time required to optimize the parameters of the model while maintaining the prediction accuracy and has a simpler structure and faster running speed. Considering the timeliness of traffic checkpoint data and the large amount of vehicle trajectory sequence datavely obtain the time-varying features of vehicle trajectory data, we adopted the GRU network cell model to extract the temporal features.

Regarding time-series features, the output of the multi-head self-attention serves as the input, employing the gated fusion mechanism of the GRU model for the integration. During this process, a gating mechanism dynamically adjusts the information weight at each time step, enhancing the capture of evolving patterns in the vehicle trajectory data. Finally, the temporal characteristics of the data were mined through information transfer among the units in the GRU model. The operations within the temporal encoder are defined as follows:
\begin{equation}
    u_t=\sigma\left(W_u\left[\text{MHSelfAttention}(Q,K,V), h_{t-1}\right]+b_u\right),
\end{equation}
\begin{equation}
    r_t=\sigma\left(W_r\left[\text{MHSelfAttention}(Q,K,V), h_{t-1}\right]+b_r\right) ,
\end{equation}
\begin{equation}
    c_t=\tanh \left(W_c\left[\text{MHSelfAttention}(Q,K,V),\left(r_t * h_{t-1}\right)\right]+b_c\right) ,
\end{equation}
\begin{equation}
    b_t=u_t * h_{t-1}+\left(1-u_t\right) * c_t,
\end{equation}where $u_t$ represents the update gate, $r_t$ represents the reset gate, $c_t$ represents the candidate data. $W_r$ and $W_c$ are weight matrices, and $b_u$, $b_r$ and $b_c$ are bias terms.

\subsection{Decoder}
As shown in Fig. \ref{f2}, the $b_t$ obtained from the temporal encoder is fed into a regression layer. The regression layer uses a fully connected layer to further process the representations output by the temporal encoder to obtain the final forecasting results. The operations within the decoder are defined as follows:
\begin{equation}
    Y=\text{tanh}(\text{FC}(b_t+\sigma(\text{RAGCN}(X,A_f))+\text{Conv1d}(\mathcal{X}))),
\end{equation}where $\sigma$ represents the sigmoid function, tanh is the activation function, and $Y \in R^{N\times T}$ denotes the traffic signal to be predicted. We employed a non-autoregressive approach to generate forecasting results to enhance computational efficiency and minimize error accumulation.

\section{EXPERIMENT RESULTS AND ANALYSIS}
In this section, we describe extensive experiments conducted on three real-world datasets to address the following six research questions\footnote{The implementation will be publicly accessible at \url{https://github.com/Xinyu-2003/PSIRAGCN} once this manuscript is accepted.}
:

\begin{itemize}
    \item \textbf{RQ1:} How does our proposed PSIRAGCN perform compared with the various State-of-the-art baselines?
    \item \textbf{RQ2:} How do the different components of PSIRAGCN affect its performance?
    \item \textbf{RQ3:} How do hyperparameters affect PSIRAGCN?
    \item \textbf{RQ4:} How does the efficiency of PSIRAGCN compare to the baselines?
\end{itemize}

\subsection{Experimental Settings}
\textit{\textbf{1) Datasets:}} We conduct extensive experiments on eight real-world traffic datasets, which include three highway traffic flow datasets \cite{39} (PEMS04, PEMS07, PEMS08). Table \ref{t2} provides detailed statistics for these datasets. The highway traffic flow datasets were collected by Caltrans’s Performance Measurement System (PEMS) \cite{52} and aggregated into 5-minute observations.
\begin{table}[!htbp]
\centering
\captionsetup{font=footnotesize, justification=centering, skip=1pt, belowskip=-6pt}
\caption{\\STATISTICS OF THE EIGHT TRAFFIC DATASETS}
\label{t2}
\renewcommand{\arraystretch}{1.5}
\begin{adjustbox}{width=\columnwidth} 
\centering
    \begin{tabular}{c|c|c|c|c}
    \hline
        Dataset & Nodes & Granularity & Samples & Time range \\ \hline \hline
        PEMS04 & 307 & 5min & 16992 & 01/01/2018-02/28/2018 \\ 
        PEMS07 & 883 & 5min & 28224 & 05/01/2017-08/31/2017 \\ 
        PEMS08 & 170 & 5min & 17856 & 07/01/2016-08/31/2016  \\ \hline
\end{tabular}
\end{adjustbox}
\end{table}

On the highway traffic flow datasets and the traffic demand datasets, we use the data from the past 6 time horizons to predict the data from the next 6 time horizons (multistep forecasting) and split them into training, validation, and test sets in the ratio of 6:2:2. Additionally, we used Z-score normalization on all datasets to standardize the inputs.

\textit{\textbf{2) Model Settings:}} Experiments are conducted under a computer environment with one 13th Gen Intel (R) Core (TM) i5-13450HX 2.40GHz processor and an NVIDIA 4050 GPU graphics card. We trained our model using the Adam optimizer \cite{54}, and the initial learning rate was set to 0.001. The batch size was set to 64 for the highway traffic flow datasets. The number of training epochs was set to 200, and an early stop mechanism was used during training to ensure that the model was over-optimized.

\textit{\textbf{3) Evaluation Metrics:}} To evaluate the models’ performance, we use the following evaluation metrics in the experiments: Mean Absolute Error (MAE), Mean Absolute Percentage Error (MAPE), and Root Mean Square Error (RMSE). We selected the MAE as the loss function. Consistent with a previous study \cite{9}, missing values in the highway traffic flow and traffic demand datasets were masked during both the training and testing phases. Additionally, samples with flow values below 10 for the grid-based urban traffic dataset were masked to ensure consistency with previous studies \cite{53}.

\subsection{Baseline Models:}
We compared PSIRAGCN with the following baselines:

\textit{\textbf{1) Traditional Models:}}
\begin{itemize}
    \item HA: Historical Average uses the average results of historical data to predict future data. 
    \item VAR \cite{55}: Vector Auto-Regression is a time series model that captures traffic data’s temporal correlations.
    \item SVR \cite{28}: Support Vector Regression uses support vector machines to do regression on traffic sequences.
    \item LSTM \cite{18}: The Long Short-Term Memory network, a neural network-based model, effectively captures temporal correlations in time-series data. 
    \item TCN \cite{8}: The Temporal Convolutional Neural Network is implemented through the stacking of causal dilation convolutions, efficiently capturing temporal correlations.
\end{itemize}

\textit{\textbf{2) Models Designed for Graph-Based Datasets:}}
\begin{itemize}
    \item DCRNN \cite{6}: This model is an encoder-decoder structure that combines diffusion GCN with GRU to capture traffic data’s spatial-temporal correlations. 
    \item STGCN \cite{14}: This model combines the spectral GCN with 1D convolution to capture spatial-temporal correlations.
    \item ASTGCN \cite{38}: This model captures spatial-temporal correlations by designing spatial and temporal attention mechanisms, respectively.
    \item GWN \cite{9}: This model combines gated TCN with spatial GCN and proposes an adaptive adjacency matrix to learn dynamic spatial correlations.
    \item AGCRN \cite{23}: This model learns spatial-temporal correlations through the mix-hop propagation layer in the spatial module, the dilated inception layer in the temporal module, and a more refined graph learning layer. 
    \item ASTGNN \cite{12}: This self-attention-based traffic forecasting model combines a time-trending self-attention mechanism with a defined dynamic GCN.
\end{itemize}

\subsection{Comparison and Analysis of Results (RQ1):}
\textit{\textbf{1) Overall Comparison:}} The experimental results on the three types of datasets are shown in Table \ref{t3}, respectively. For all the traffic datasets, the results are the average MAE, RMSE, and MAPE values for the prediction horizons. The experimental results show that the comprehensive performance of PSIRAGCN outperforms the baselines on all eight flow datasets, particularly on the PEMS08. Our proposed model demonstrates stable and excellent performance across various forecasting scenarios, demonstrating its robust capability to extract spatiotemporal representations.

\begin{table*}[!htbp]
\centering
\captionsetup{font=normal, justification=centering, skip=1pt, belowskip=-6pt}
\caption{\\PERFORMANCE ON HIGHWAY TRAFFIC FLOW DATASETS. \textbf{BOLD}: BEST, \underline{UNDERLINE}: SECOND BEST}
\label{t3}
\renewcommand{\arraystretch}{1.5}
% \begin{adjustbox}{width=2\columnwidth} 
    \centering
    \begin{tabular}{c|c|c|c|c|c|c|c|c|c}
    \hline
        \multirow{2}{*}{Models} & \multicolumn{3}{c|}{PEMS04}& \multicolumn{3}{c|}{PEMS08}& \multicolumn{3}{c}{PEMS07} \\ \cline{2-10}
         & MAE & RMSE & MAPE & MAE & RMSE & MAPE & MAE & RMSE & MAPE \\ \hline \hline
       HA & 20.688 & 31.908 & 0.093 & 16.747 & 25.428 & 0.056 & 23.155 & 34.717 & 0.060 \\ 
VAR & 24.365 & 37.997 & 0.080 & 18.274 & 27.071 & 0.056 & 42.920 & 65.627 & 0.094 \\ 
LSTM & 22.190 & 36.971 & 0.080 & 19.510 & 29.517 & 0.059 & 30.528 & 53.123 & 0.066 \\ 
TCN & 18.119 & 28.846 & 0.067 & 14.515 & 22.000 & \textbf{0.042} & 19.065 & 29.419 & 0.045 \\ \hline
DCRNN & 18.509 & 29.375 & 0.064 & 14.932 & 22.904 & 0.050 & 19.823 & 30.374 & 0.044 \\ 
STGCN & 17.746 & 28.486 & 0.065 & 13.626 & 21.106 & 0.047 & 18.539 & 28.938 & 0.043 \\ 
Graph WaveNet & 17.586 & 28.150 & 0.082 & 13.505 & 20.829 & 0.047 & 18.768 & \underline{28.906} & 0.046 \\ 
ASTGCN & 17.489 & 28.065 & \underline{0.063} & 13.710 & 20.933 & 0.047 & 22.183 & 34.421 & 0.056 \\ 
AGCRN & \textbf{17.416} & \textbf{28.008} & \textbf{0.051} & 14.680 & 22.246 & 0.048 & \underline{18.070} & 28.212 & \underline{0.041} \\ 
ASTGNN & 17.432 & 28.040 & 0.067 & 13.870 & \underline{20.742} & 0.045 & 19.067 & 30.699 & 0.050 \\ 
\textbf{PSIRAGCN} & \underline{17.428} & \underline{28.040} & 0.077 & \textbf{13.453} & \textbf{20.502} & \underline{0.044} & \textbf{18.045} & \textbf{27.134} & \textbf{0.041} \\ \hline
    \end{tabular}
% \end{adjustbox}
\end{table*}

The results in Table \ref{t3} indicate that the traditional models HA, VAR, and SVR do not yield satisfactory performances. Although these models account for temporal correlations, they overlook the intricate spatial correlations in traffic data. The earliest proposed STGNNs, such as the DCRNN and STGCN, outperform traditional models. This is attributed to their consideration of spatial correlation modeling and pioneering the employment of predefined graph structures to enhance the capture of spatial correlations. The subsequent introduction of adaptive graph structure methods, such as GWN and AGCRN, improved the performance by utilizing trainable adaptive embeddings to model dynamic spatial correlations. However, these models still describe static spatial correlations because the generation of graph structures cannot dynamically adapt as the input data change, such as the DGCRN, which outperforms models that statistically model spatial correlations. As evidenced by the results in Table \ref{t3}, dynamic modeling of spatial associations is necessary to ensure a competitive model.

Notably, models based on the attention mechanism, such as the ASTGNN, also demonstrated strong performance. This is attributed to the effective calculation of the correlation degrees between different time steps and nodes by the attention mechanism, allowing for a better capture of spatial-temporal dependencies over both short and long ranges. 

Our proposed PSIRAGCN exhibits superior performance for the following reasons: \textit{i)} The PSIRAGCN uses a pattern-spatial interactive fusion strategy to share spatial-pattern information between input sequence data and capture spatial pattern correlations from global to local change granularities. In the spatial-temporal interactive learning process, the captured temporal correlations and the captured spatial correlations positively affect each other. This feedback mechanism allows PSIRAGCN to explore deeper spatial-pattern correlations. \textit{ii)} By fully leveraging spatial-pattern information, we employ a dynamic regional massage passing method to simulate the dynamic regional regional heterogeneity between nodes in the traffic network. Simultaneously, we used a regional characteristics bank to store the valuable regional heterogeneity for each node, mitigating the impact of spatial heterogeneity. 

\subsection{Ablation Study (RQ2)} To further verify the effectiveness of each component in PSIRAGCN, we conduct an ablation study on the PEMS04 and PEMS08. We designed four variants of the PSIRAGCN as follows: 
\begin{itemize}
    \item \textbf{w/o PSI:} PSIRAGCN removes the PSI module;
    \item \textbf{w/o RAGCN:} PSIRAGCN replaces the RAGCN with a normal GCN; 
    \item \textbf{w/o RCB:} PSIRAGCN removes the regional characteristics bank from the RAGCN; 
    \item \textbf{w/o PEConv:} PSIRAGCN removes PEConv in PSF module; 
\end{itemize}

\begin{figure*}
    \centering
        \centering
        \includegraphics[width=\linewidth]{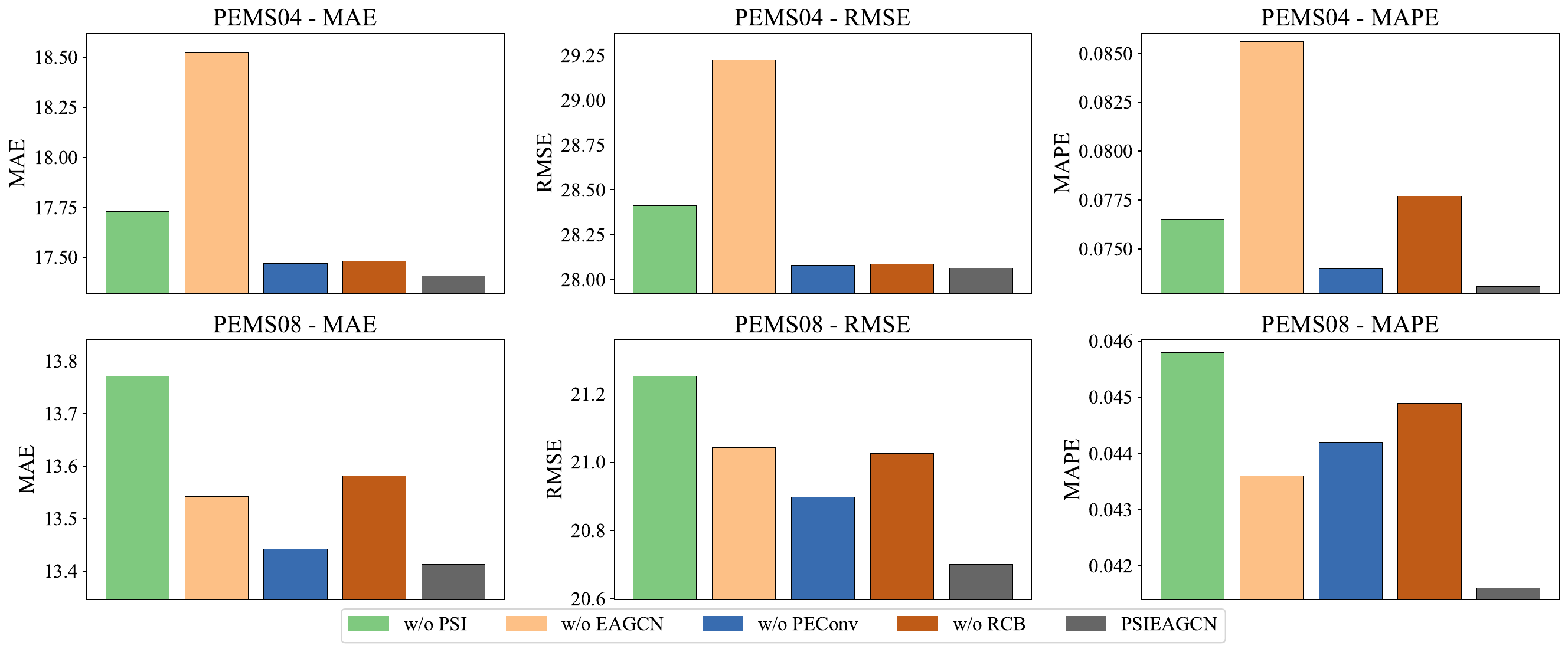}
        \captionsetup{font=footnotesize, skip=1pt}
        \caption{Ablation study on two traffic datasets.}
        \label{f5}
\end{figure*}

The results of the ablation experiments are shown in Fig. \ref{f5}. The effectiveness of the different components in the PSIRAGCN has similar distributions on the experimental datasets. The component with the most significant impact on model performance was the RAGCN. The RAGCN utilizes its internally generated dynamic regional message passing for spatial representation mining, endowing the PSIRAGCN with the ability to capture dynamic regional spatial correlations. The result of \textbf{w/o RAGCN} indicates that the performance of PSIRAGCN significantly deteriorates when RAGCN is omitted, as it loses its spatial modeling capability. Similarly, the result of \textbf{w/o RB} suggests that although PSIRAGCN possesses spatial modeling capabilities, the absence of regional message passing and the failure to consider the distinct regional heterogeneity at each node limit its ability to capture regional spatial correlations. Moreover, the result of \textbf{w/o PEConv } demonstrates that if the model cannot capture pattern features and only engages in the interactive capture of spatial correlations, the model performance is compromised. This is because spatial-pattern interactive fusion results in positive feedback between the captured spatial and pattern features. The model naturally exhibits poorer performance when this positive feedback in both the spatial and temporal dimensions is lost. Furthermore, the result of \textbf{w/o PSF:} indicates that the interactive fusion between spatial and pattern features is more effective than the paradigm of just capturing spatial correlations. This is because interactive fusion enables mutual reinforcement and positive feedback between spatial and pattern features.

\subsection{Parameter Sensitivity Analysis (RQ3)}
\begin{figure}
    \centering
        \centering
        \includegraphics[width=\linewidth]{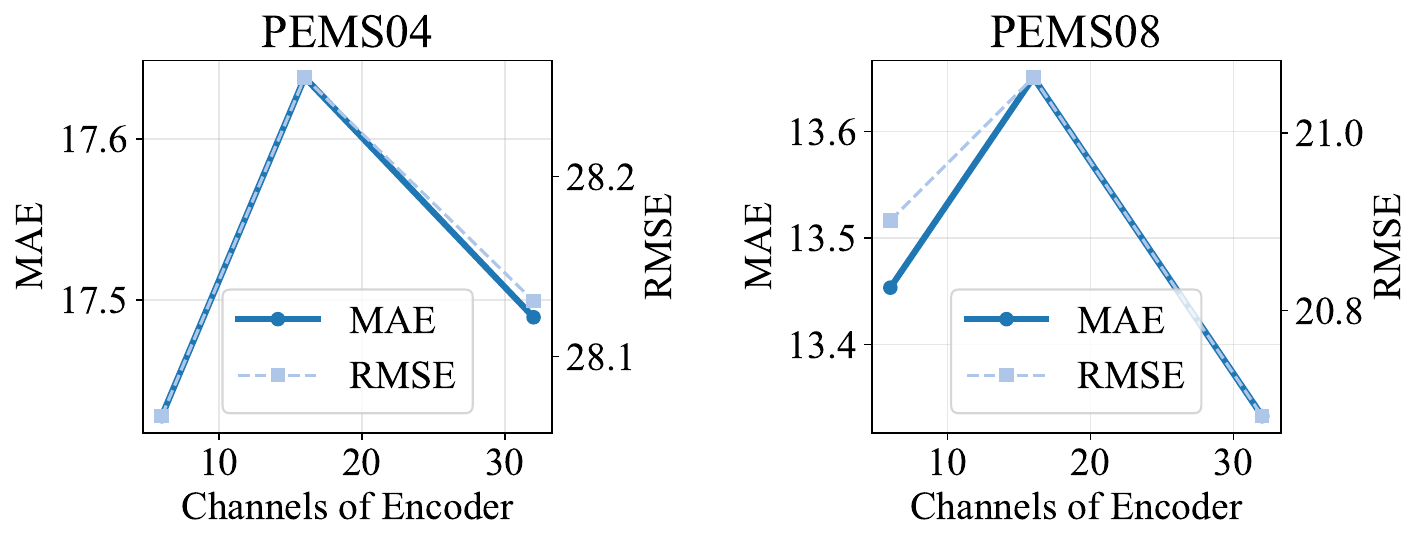}
        \captionsetup{font=footnotesize, skip=2pt}
        \caption{Hyperparameter study on two traffic datasets.}
        \label{f7}
\end{figure}

To further investigate the parameter sensitivity, we conducted hyperparameter studies on PEMS04 and PEMS08. We selected the number of feature channels in the encoder to investigate their impact on the model performance. The experimental results are shown in Fig. \ref{f7}. First, as the number of feature channels increased, the performance of the PSIRAGCN gradually improved because having more feature channels allowed the PSIRAGCN to learn higher-dimensional spatiotemporal features. However, when the number of feature channels reaches a certain threshold, the performance of PSIRAGCN stabilizes or even declines. The detailed optimal hyperparameter settings for the PSIRAGCN on each dataset are listed in Table \ref{t4}.

\begin{table}[!ht]
    \centering
    \captionsetup{font=scriptsize, justification=centering, skip=1pt, belowskip=-6pt}
    \caption{\\HYPERPARAMETER SETTINGS}
    \label{t4}
    \renewcommand{\arraystretch}{1.5}
    \begin{adjustbox}{width=\columnwidth} 
    \begin{tabular}{c|c|c|c|c|c|c|c}
    \hline
        \multirow{2}{*}{Dataset} & Batch & Learning & Weight & Epoch & Channels & Kernel Size & Diffusion \\ 
         & Size & Rate & Decay &   & $C$ & $s1,s2$ & Step $k$
        \\
        \hline \hline
        PEMS04 & 64 & 0.001 & 0.0001 & 300 & 6 & [6, 1] & 2  \\ 
        PEMS07 & 64 & 0.001 & 0.0001 & 300 & 6 & [6, 1] & 2  \\ 
        PEMS08 & 64 & 0.001 & 0.0001 & 300 & 6 & [6, 1] & 2 \\ \hline
      \end{tabular}
    \end{adjustbox}
\end{table}

\subsection{Efficiency Study (RQ4)}
\begin{figure}
    \centering
        \centering
        \includegraphics[width=\linewidth]{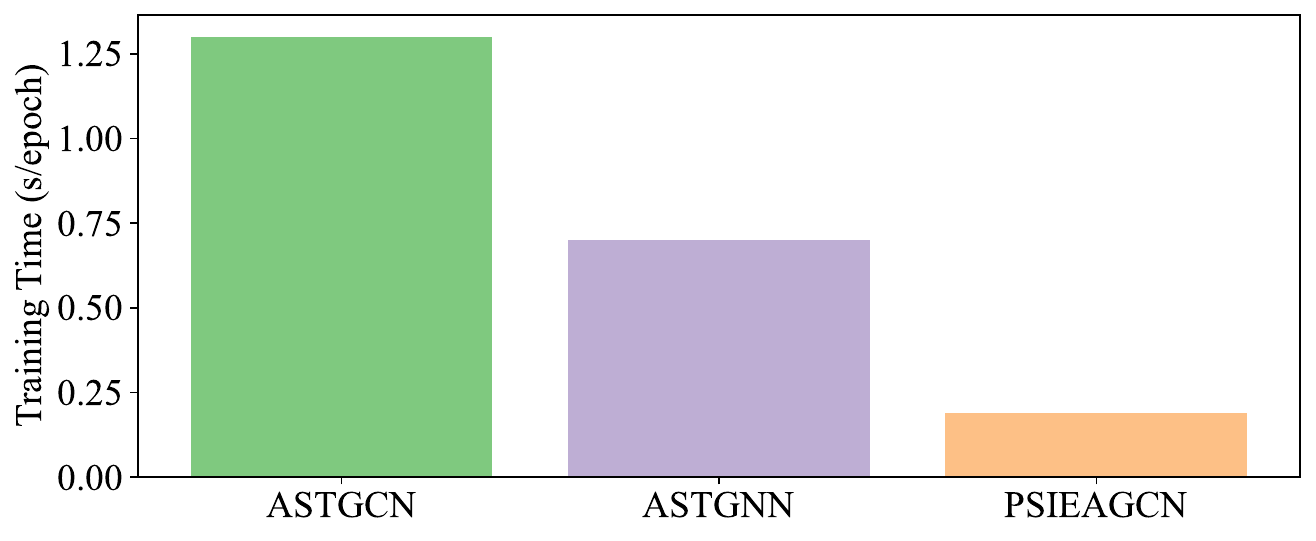}
        \captionsetup{font=footnotesize, skip=2pt}
        \caption{Computational costs on PEMS04.}
        \label{f8}
\end{figure}

In this section, we compare PSIRAGCN with other models on the PEMS04 and PEMS08 datasets in terms of computational costs. As shown in Fig. \ref{f8}, we compare training time. All models were run in the same experimental environment, and the batch size was uniformly set to 64. Models such as ASTGNN, which are based on attention mechanisms, exhibit excellent performance; however, their frequent attention calculations lead to significant memory usage. The PSIRAGCN offers excellent performance without incurring significantly higher computational costs compared to these State-of-the-art baselines.

\section{CONCLUSION}
In this study, we proposed a Pattern-Spatial Interactive Fusion and Regional Awareness Graph Convolutional Network (PSIRAGCN) for traffic forecasting. We proposed a pattern-spatial interactive fusion strategy to capture relationships by incorporating feature interactions between pattern and spatial information. Additionally, we introduced an effective RAGCN module based on the message-passing method for modeling spatial correlations. This module fully utilized the traffic net and incorporates distinct regional heterogeneity at each node. The effective capture of these spatial correlations was then propagated through a pattern-spatial interactive fusion strategy to reveal deeper patterns and spatial associations in the data. This process established a pattern-spatial feedback mechanism that optimizes the prediction results. Extensive experiments on three real-world datasets demonstrated that our model outperformed the comparison baselines while achieving a balance between computational costs. Nevertheless, PSIRAGCN remained a non-lightweight model. It may requires substantial computational resources when deployed on large-scale traffic networks with tens of thousands of nodes. Therefore, investigating methods to reduce the computational resources required for spatial-temporal interactive learning is a promising avenue for further research.

\bibliographystyle{IEEEtran}  % 使用 IEEEtran 样式
\bibliography{paper}     % references 是你的 .bib 文件名（不带 .bib 扩展名）

% \newpage

 

% \end{IEEEbiography}


% % \bf{If you will not include a photo:}\vspace{-33pt}
% % \begin{IEEEbiographynophoto}{John Doe}
% % Use $\backslash${\tt{begin\{IEEEbiographynophoto\}}} and the author name as the argument followed by the biography text.
% % \end{IEEEbiographynophoto}

\vfill

\end{document}